\pdfoutput=1

\documentclass[letterpaper, 10 pt, conference]{ieeeconf}  

\IEEEoverridecommandlockouts                              

\usepackage{graphicx}        
\usepackage{multicol}        
\usepackage[bottom,hang,flushmargin]{footmisc}
\usepackage{todonotes}
\usepackage{algorithm}
\usepackage{algorithmic}
\usepackage{siunitx}
\usepackage{float}
\usepackage{amsmath}
\usepackage{amsfonts}
\usepackage{amssymb}
\usepackage{subfigure}

\newcommand{\etal}{\textit{et al.}}
\DeclareMathOperator*{\argmax}{argmax}

\title{\LARGE \bf
Zero-Shot Skill Composition and Simulation-to-Real Transfer by Learning Task Representations
}

\author{
Zhanpeng He\thanks{$^*$ These authors contributed equally to the paper.}$^*$, Ryan Julian$^*$, Eric Heiden, Hejia Zhang, \\Stefan Schaal, Joseph J. Lim, Gaurav Sukhatme, and Karol Hausman
\thanks{\noindent Zhanpeng He, Ryan Julian, Eric Heiden, Hejia Zhang, Stefan Schaal, Joseph J. Lim and Gaurav Sukhatme are with the Department of
Computer Science, University of Southern California, Los Angeles, USA.
        {\tt \{zhanpenh, rjulian, heiden, hejia, sschaal, limjj, gaurav\}@usc.edu}}%
\thanks{\noindent Karol Hausman is with Google Brain, Mountain View, CA, USA
        {\tt karolhausman@google.com}}
}

\begin{document}

\maketitle
\thispagestyle{empty}
\pagestyle{empty}

\begin{abstract}
Sim-to-real transfer is an important strategy for making reinforcement learning practical with real robots. Successful sim-to-real transfer systems have difficulty producing policies which generalize across tasks, despite training for thousands of hours equivalent real robot time. To address this shortcoming, we present a novel approach to efficiently learning new robotic skills directly on a real robot, based on model-predictive control (MPC) and an algorithm for learning task representations. In short, we show how to reuse the simulation from the pre-training step of sim-to-real methods as a tool for foresight, allowing the sim-to-real policy adapt to unseen tasks. Rather than end-to-end learning policies for single tasks and attempting to transfer them, we first use simulation to simultaneously learn (1) a continuous parameterization (i.e. a skill embedding or latent) of task-appropriate primitive skills, and (2) a single policy for these skills which is conditioned on this representation. We then directly transfer our multi-skill policy to a real robot, and actuate the robot by choosing sequences of skill latents which actuate the policy, with each latent corresponding to a pre-learned primitive skill controller. We complete unseen tasks by choosing new sequences of skill latents to control the robot using MPC, where our MPC model is composed of the pre-trained skill policy executed in the simulation environment, run in parallel with the real robot. We discuss the background and principles of our method, detail its practical implementation, and evaluate its performance by using our method to train a real Sawyer Robot to achieve motion tasks such as drawing and block pushing.
\end{abstract}

\section{INTRODUCTION}
Reinforcement learning algorithms have been proven to be effective to learn complex skills in simulation environments in \cite{mnih2015dqn}, \cite{DBLP:journals/nature/SilverHMGSDSAPL16} and \cite{DBLP:journals/corr/YahyaLKCL16}. However, practical robotic reinforcement learning for complex motion skills remains a challenging and unsolved problem, due to the high number of samples needed to train most algorithms and the expense of obtaining those samples from real robots. Most existing approaches to robotic reinforcement learning either fail to generalize between different tasks and among variations of single tasks, or only generalize by requiring collecting impractical amounts of real robot experience. With recent advancements in robotic simulation, and the widespread availability of large computational resources, a popular family of methods seeking to address this challenge has emerged, known as ``sim-to-real'' methods. These methods seek to offload most training time from real robots to offline simulations, which are trivially parallelizable and much cheaper to operate. Our method combines this ``sim-to-real'' schema with representation learning and model-predictive control (MPC) to make transfer more robust, and to significantly decrease the number of simulation samples needed to train policies which achieve families of related tasks.
\begin{figure}
    \centering
    \begin{subfigure}
        \centering
        \includegraphics[width=4cm]{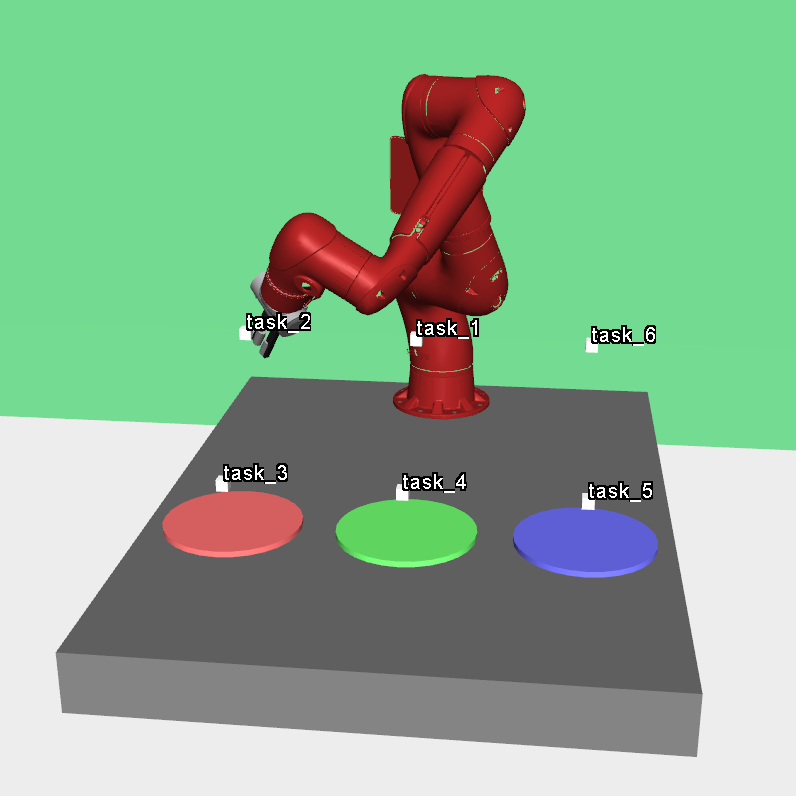}
        \label{fig:reach_sim}
    \end{subfigure}
    \begin{subfigure}
        \centering
        \includegraphics[width=4cm]{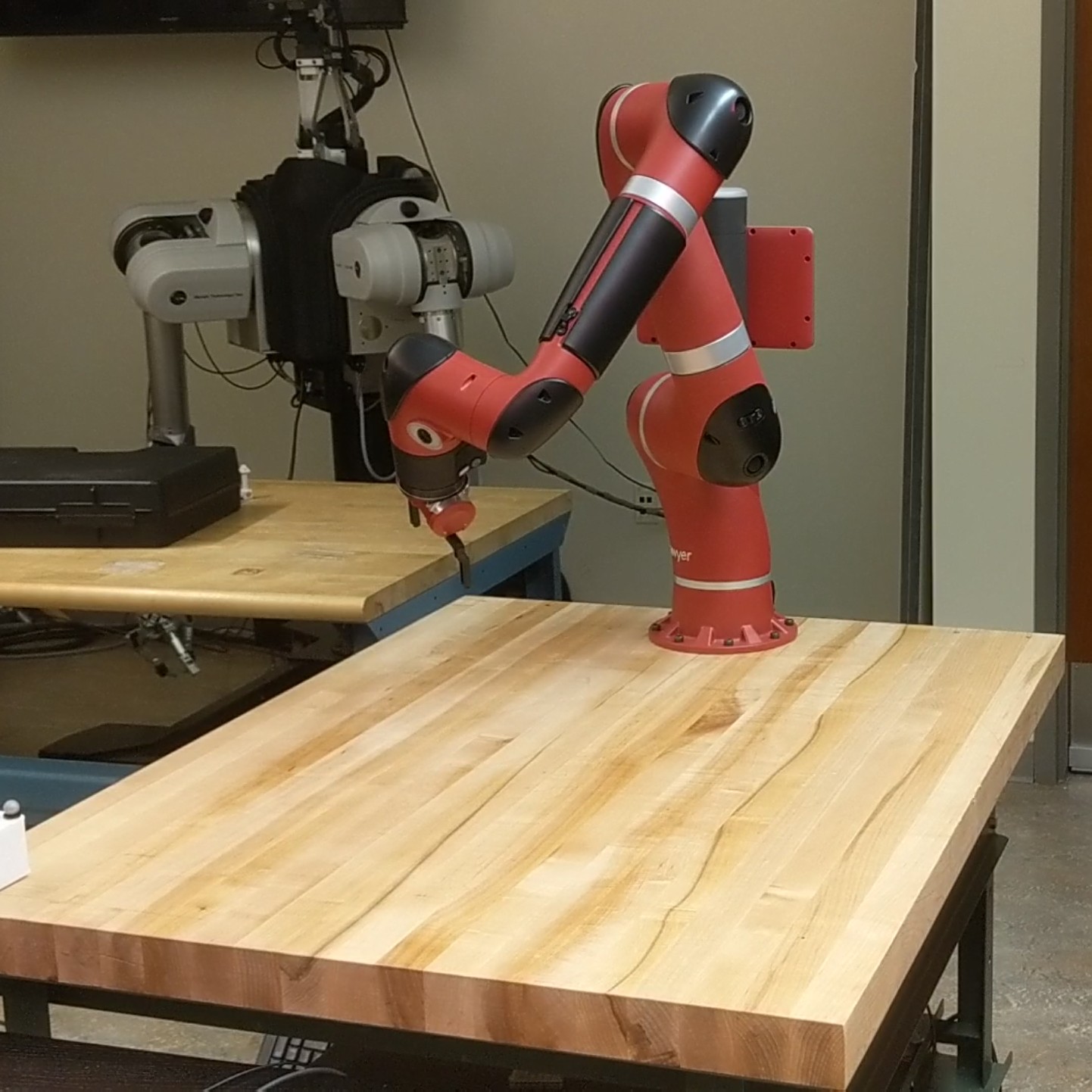}
        \label{fig:reach}
    \end{subfigure}
    \caption{The Sawyer robot performing the reaching task in simulation (left) and real world (right)}
    \label{fig:robots}
\end{figure}

\begin{figure}
    \centering
    \begin{subfigure}
        \centering
        \includegraphics[width=4cm]{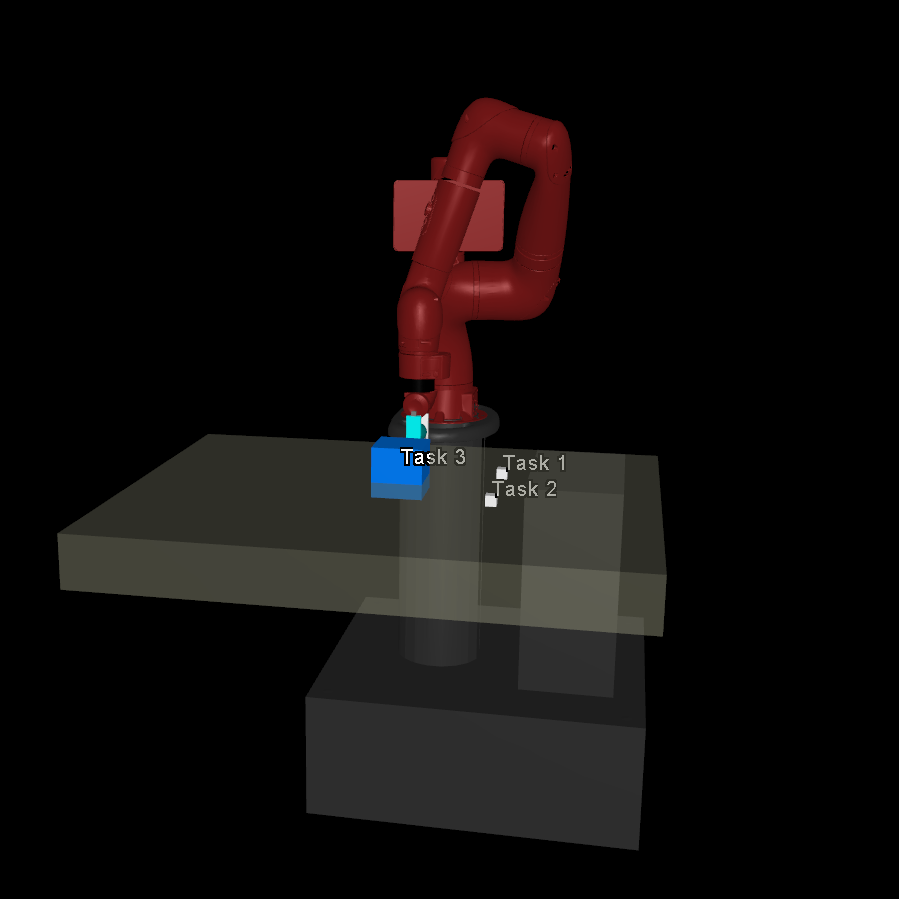}
        \label{fig:push_sim}
    \end{subfigure}
    \begin{subfigure}
        \centering
        \includegraphics[width=4cm]{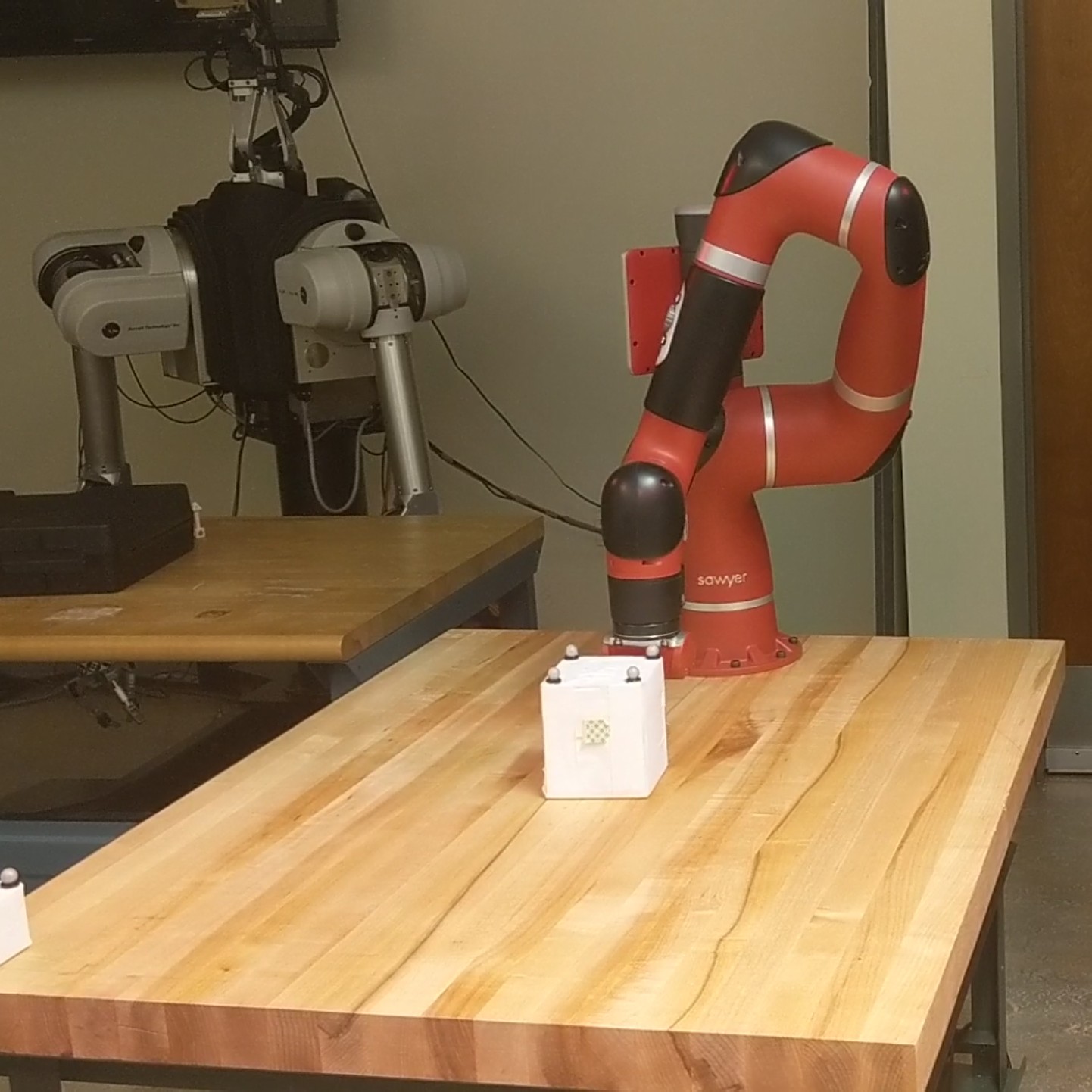}
        \label{fig:push}
    \end{subfigure}
    \caption{The Sawyer robot performing the box pushing task in simulation (left) and real world (right)}
    \label{fig:robots}
\end{figure}
The key insight behind our method is that the simulation used in the pre-training step of a simulation-to-real method can also be used online as a tool for foresight. It allows us to predict the behavior of a known policy on an unseen task. When combined with a latent-conditioned policy, where the latent actuates variations of useful policy behavior (e.g. skills), this simulation-as-foresight tool allows our method to use what the robot has already learned to do (e.g. the pre-trained policy) to bootstrap online policies for tasks it has never seen before. That is, given a latent space of useful behaviors, and a simulation which predicts the rewards for those behaviors on a new task, we can reduce the adaptation problem to intelligently choosing a sequence of latent skills which maximize rewards for the new task.

\section{BACKGROUND}
Most simulation-to-real approaches so far have focused on addressing the ``reality gap'' problem. The reality gap problem is the domain shift performance loss induced by differences in dynamics and perception between the simulation (policy training) and real (policy execution) environments. Training a policy only in a flawed simulation generally yields control behavior which is not adaptable to even small variations in the environment dynamics. Furthermore, simulating the physics behind many practical robotic problems (e.g. sliding friction and contact forces) is an open problem in applied mathematics, meaning it is not possible to solve a completely accurate simulation for many important robotic tasks \cite{10.1007/3-540-59496-5_337}. Rather than attempt to create an explicit alignment between simulation and real \cite{closingrealitygap}, or randomize our simulation training to a sufficient degree to learn a policy which generalizes to nearby dynamics \cite{peng2017simreal}, our method seeks to learn a sufficient policy in simulation, and adapt it quickly to the real world online during real robot execution.

Our proposed approach is based on four key components: reinforcement learning with policy gradients (RL) \cite{DBLP:journals/corr/SchulmanAC17}, variational inference \cite{DBLP:journals/corr/KingmaW13}, model-predictive control (MPC), and physics simulation. We use variational inference to learn a low-dimensional latent space of skills which are useful for tasks, and RL to simultaneously learn single policy which is conditioned on these latent skills. The precise long-horizon behavior of the policy for a given latent skill is difficult to predict, so we use MPC and an online simulation to evaluate latent skill plans in the in simulation before executing them on the real robot.

\section{RELATED WORK}
Learning skill representations to aid in generalization has been proposed in works old and new. Previous works proposed frameworks such as Associative Skill Memories~\cite{pastor2012asm} and probabilistic movement primitives~\cite{rueckert2015movprim} to acquire a set of reusable skills. Our approach is built upon \cite{hausman2018learning}, which learns a embedding space of skills with reinforcement learning and variational inference, and \cite{julian2018scaling} which shows that these learned skills are transferable and composable on real robots. While \cite{julian2018scaling} noted that predicting the behavior of latent skills is an obstacle to using this method, our approach addresses the problem by using model-predictive control to successfully complete unseen tasks with no fine-tuning on the real robot. Exploration is a key problem in robot learning, and our method uses latent skill representations to address this problem. Using learned latent spaces to make exploration more tractable is also studied in \cite{gu2017deep} and \cite{eysenbach2018diversity}. Our method exploits a latent space for task-oriented exploration: it uses model-predictive control and simulation to choose latent skills which are locally-optimal for completing unseen tasks, then executes those latent skills on the real robot.

Using reinforcement learning with model-predictive control has been explored previously. Kamthe \etal~\cite{DBLP:journals/corr/KamtheD17} proposed using MPC to increase the data efficiency of reinforcement algorithms by training probabilistic transition models for planning. In our work, we take a different approach by exploiting our learned latent space and simulation directly to find policies for novel tasks online, rather than learning and then solving a model.

Simulation-to-real transfer learning approaches include randomizing the dynamic parameters of the simulation~\cite{peng2017simreal}, and varying the visual appearance of the environment~\cite{sadeghi2017cadrl}, both of which scale linearly or quadratically the amount of computation needed to learn a transfer policy. Other strategies, such as that of Barrett \etal~\cite{AAMASWS10-barrett} reuse models trained in simulation to make sim-to-real transfer more efficient, similar to our method, however this work requires an explicit pre-defined mapping between seen and unseen tasks. Saemundson \etal~\cite{DBLP:journals/corr/abs-1803-07551} use meta-learning and learned representations to generalize from pre-trained seen tasks to unseen tasks, however their approach requires that the unseen tasks be very similar to the pre-trained tasks, and is few-shot rather than zero-shot. Our method is zero-shot with respect to real environment samples, and can be used to learn unseen tasks which are significantly out-of-distribution, as well as for composing learned skills in the time domain to achieve unseen tasks which are more complex than the underlying pre-trained task set.

Our work is closely related to simultaneous work performed by Co-Reyes \etal~\cite{coreyes2018self}. Whereas our method learns an explicit skill representations using pre-chosen skills identified by a known ID, \cite{coreyes2018self} learn an implicit skill representation by clustering trajectories of states and rewards in a latent space. Furthermore, we focus on MPC-based planning in the latent space to achieve robotic tasks learned online with a real robot, while their analysis focuses on the machine learning behind this family of methods and uses simulation experiments.

\section{METHOD}

\subsection{Skill Embedding Algorithm}
\label{sec:skill_embedding_algorithm}
In our multi-task RL setting, we pre-define a set of low-level skills with IDs $\mathcal{T} = \{1,\ldots,N\}$, and accompanying, per-skill reward functions $r^t(s, a)$.

In parallel with learning the joint low-level skill policy $\pi_\theta$ as in conventional RL, we learn an embedding function $p_\phi$ which parameterizes the low-level skill library using a latent variable $z$. Note that the true skill identity $t$ is hidden from the policy behind the embedding function $p_\phi$.
Rather than reveal the skill ID to the policy, once per rollout we feed the skill ID $t$, encoded as s one-hot vector, through the stochastic embedding function $p_\phi$ to produce a latent vector $z$. We feed this same value of $z$ to the policy for the entire rollout, so that all steps in a trajectory are correlated with the same value of $z$.
\begin{align}
    \label{eq:objective}
    \mathfrak{L}(\theta, \phi, \psi) &=\nonumber\\
    \max_\pi \:&\mathbb{E}_{\substack{\pi(a, z|s, t) \\ t\in T}}
        \left[
            \sum_{i=0}^\infty \gamma^i \hat{r}(s_i, a_i, z, t) \bigg| s_{i+1}
        \right]
    \nonumber
    \\[-3em]
    &\intertext{where} \nonumber\\[-2.5em]
    \nonumber
    \hat{r}(s_i, a_i, z, t) &= \alpha_1 \mathbb{E}_{t\in T} \left[\mathbb{H}\left(p_\phi(z|t)\right)\right] + \alpha_2 \log q_\psi(z | s_i^H) \nonumber \\
    \nonumber
    &~~~ + \alpha_3 \mathbb{H}\left(\pi_\theta(a_i | s_i, z)\right) + r^t(s_i, a_i)
    \nonumber
\end{align}

To aid in learning the embedding function, we learn an inference function $q_\psi$ which, given a state-only trajectory window $s_i^H$ of length $H$, predicts the latent vector $z$ which was fed to the low-level skill policy when it produced that trajectory. This allows us to define an augmented reward which encourages the policy to produce distinct trajectories for different latent vectors. We learn $q_\psi$ in parallel with the policy and embedding functions, as shown in Eq.~\ref{eq:objective}.

We add a policy entropy bonus $\mathbb{H}\left(\pi_\theta(a_i | s_i, z)\right)$, which ensures that the policy does not collapse to a single solution for each skill. For a detailed derivation, refer to~\cite{hausman2018learning}.

\subsection{Skill Embedding Criterion}
In order for the learned latent space to be useful for completing unseen tasks, we seek to constrain the embedding distribution to satisfy two important properties:
\begin{enumerate}
    \item \textbf{High entropy:} Each task should induce a distribution over latent vectors which is wide as possible, corresponding to many variations of a single skill.
    \item \textbf{Identifiability:} Given an arbitrary trajectory window, the inference network should be able to predict with high confidence the latent vector fed to the policy to produce that trajectory.
\end{enumerate}

When applied together, these properties ensure that during training the policy is trained to encode high-reward controllers for many parameterizations of a skill (high-entropy), while simultaneously ensuring that each of these latent parameterizations corresponds to a distinct variation of that skill. This dual constraint is the key for using model predictive control or other composing methods in the latent space as discussed in Sec.~\ref{sec:mpc_adaptation}.

\begin{figure}[H]
    \centering
    \includegraphics[width=\columnwidth]{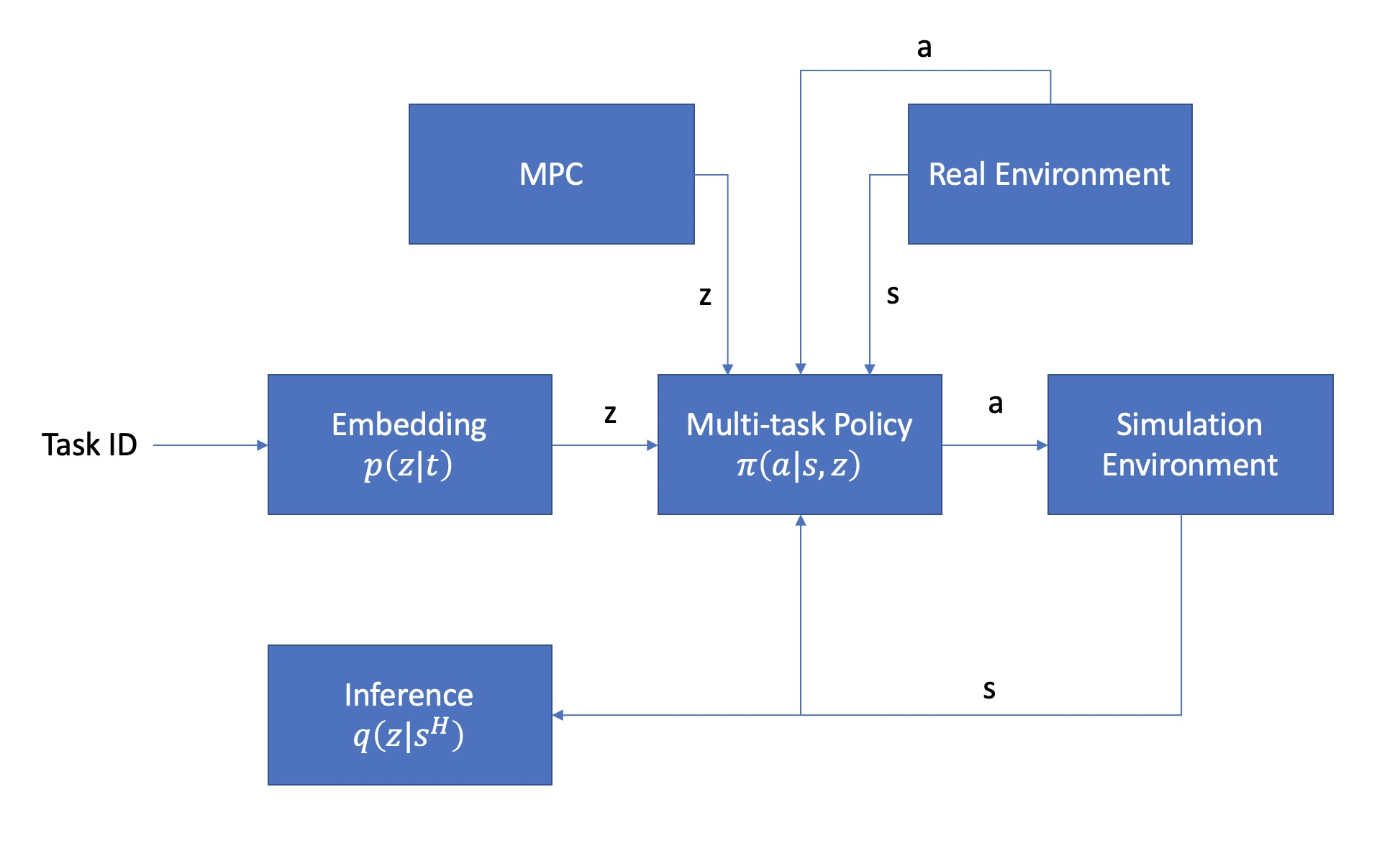}
    \caption{Skill Embedding Algorithm and MPC}
    \label{fig:mpc}
\end{figure}

We train the policy and embedding networks using Proximal Policy Optimization~\cite{schulman2017ppo}, though our method may be used by any parametric reinforcement learning algorithm. We use the MuJoCo physics engine \cite{todorov2012mujoco} to implement our Sawyer robot simulation environments. We represent the policy, embedding, and inference functions using multivariate Gaussian distributions whose mean and diagonal covariance are parameterized by the output of a multi-layer perceptron. The policy and embedding distributions are jointly optimized by the reinforcement learning algorithm, while we train the inference distribution using supervised learning and a simple cross-entropy loss.

\subsection{Using Model Predictive Control for Zero-Shot Adaptation}
\label{sec:mpc_adaptation}

To achieve unseen tasks on a real robot with no additional training, we freeze the multi-skill policy learned in Sec.~\ref{sec:skill_embedding_algorithm}, and use a new algorithm which we refer to as a ``composer.'' The composer achieves unseen tasks by choosing new sequences of latent skill vectors to feed to the frozen skill policy. Exploring in this smaller space is faster and more sample-efficient, because it encodes high-level properties of tasks and their relations. Each skill latent induces a different pre-learned behavior, and our method reduces the adaptation problem to choosing sequences of these pre-learned behaviors--continuously parameterized by the skill embedding--to achieve new tasks.

Note that we use the simulation itself to evaluate the future outcome of the next action. For each step, we set the state of the simulation environment to the observed state of the real environment. This equips our robot with with the ability to predict the behavior of different skill latents. Since our robot is trained in a simulation-to-real framework, we can reuse the simulation from the pre-training step as a tool for foresight when adapting to unseen tasks. This allow us to select a latent skill online which is locally-optimal for a task, even if that task was seen not during training. We show that this scheme allows us to perform zero-shot task execution and composition for families of related tasks. This is in contrast to existing methods, which have mostly focused on direct alignment between simulation and real, or data augmentation to generalize the policy using brute force. Despite much work on simulation-to-real methods, neither of these approaches has demonstrated the ability to provide the adaptation ability needed for general-purpose robots in the real world. We believe our method provides a third path towards simulation-to-real adaptation that warrants exploration, as a higher-level complement to these effective-but-limited existing low-level approaches.

We denote the new task $t^{\text{new}}$ corresponding to reward function $r^{\text{new}}$, the real environment in which we attempt this task $\mathcal{R}(s'|s,a)$, and the RL discount factor $\gamma$. We use the simulation environment $\mathcal{S(s'|s,a)}$, frozen skill embedding $p_\phi(z|t)$, and latent-conditioned skill policy $\pi_\theta(a|s,z)$, all trained in Sec.~\ref{sec:skill_embedding_algorithm}, to apply model-predictive control in the latent space as follows (Algorithm~\ref{algo:latent_space_mpc}).

We first sample $k$ candidate latents $\mathcal{Z} = \{z_1, \ldots, z_k\}$ according to $p(z) = \mathbb{E}_{t \sim p(t)} p_\phi(z|t)$. We observe the state $s_{\text{real}}$ of real environment $\mathcal{R}$.

For each candidate latent $z_i$, we set the initial state of the simulation $\mathcal{S}$ to $s_{\text{real}}$. For a horizon of $T$ time steps, we sample the frozen policy $\pi_\theta$, conditioned on the candidate latent $a_{j \in T} \sim \pi_\theta(a_j|s_j,z_i)$, and execute the actions $a_j$ the simulation environment $\mathcal{S}$, yielding and total discounted reward $R_i^{\text{new}} = \sum_{j=0}^T{\gamma^j r^{\text{new}}(s_j,a_j)}$ for each candidate latent. We then choose the candidate latent acquiring the highest reward $z^* = \argmax_{i} R_i^\text{new}$, and use it to condition and sample the frozen policy $a_{l \in N} \sim \pi_\theta(a_j|s_j,z^*)$ to control the real environment $\mathcal{R}$ for a horizon of $N < T$ time steps.

We repeat this MPC process to choose and execute new latents in sequence, until the task has been achieved.

\begin{figure}
    \centering
    \includegraphics[width=\columnwidth]{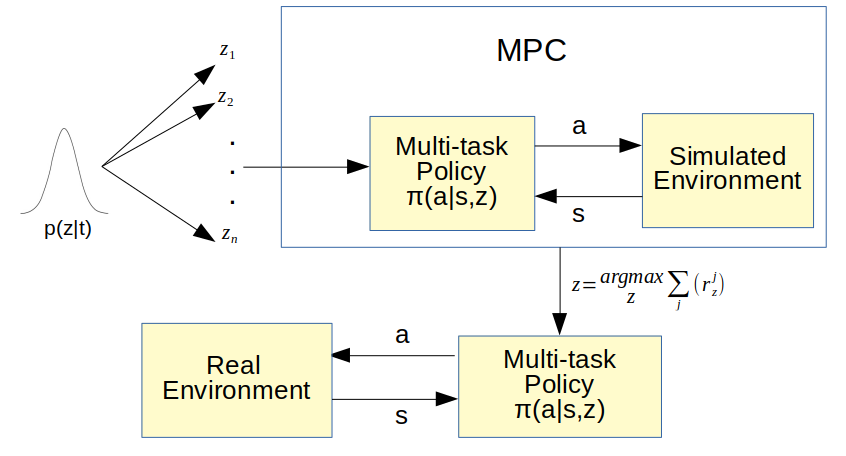}
    \caption{Using model-predictive control with embedding functions and multi-task policy}
    \label{fig:mpc}
\end{figure}

\begin{algorithm}
    \caption{MPC in Skill Latent Space}
    \begin{algorithmic}
    \label{algo:latent_space_mpc}
    \REQUIRE{A latent-conditioned policy $\pi_{\theta}(a |s, z)$, a skill embedding distribution $p_\phi(z|t)$, a skill distribution prior $p(t)$, a simulation environment $\mathcal{S}(s'|s, a)$, a real environment $\mathcal{R}(s'|s, a)$, a new task $t^{\text{new}}$ with associated reward function $r^{\text{new}}(s,a)$, an RL discount factor $\gamma$, an MPC horizon $T$, and a real environment horizon $N$.}
    \WHILE{$t^{new}$ is not complete}
    \STATE{Sample $\mathcal{Z} = \{z_1, \ldots, z_k\} \sim p(z) = \mathbb{E}_{t \sim p(t)} p_\phi(z|t)$}
    \STATE{Observe $s_{\text{real}}$ from $\mathcal{R}$}
    \FOR{$z_i \in \mathcal{Z}$}
    \STATE{Set inital state of $\mathcal{S}$ to $s_{\text{real}}$}
    \FOR{$j \in  \{1, \ldots, T\}$}
    \STATE{Sample $a_j \sim \pi_\theta(a_j|s_j,z_i)$}
    \STATE{Execute simulation $s_{j+1} = \mathcal{S}(s_j, a_j)$}
    \ENDFOR
    \STATE{Calculate $R_i^{\text{new}} = \sum_{j=0}^T{\gamma^j r^{\text{new}}(s_j,a_j)}$}
    \ENDFOR
    \STATE{Choose $z^* = \argmax_{z_i} R_i^\text{new}$}
    \FOR{$l \in \{1,\ldots, N\}$}
    \STATE{Sample $a_l \sim \pi_\theta(a_l|s_l,z^*)$}
    \STATE{Execute simulation $s_{l+1} = \mathcal{S}(s_l, a_j)$}
    \ENDFOR
    \ENDWHILE
    \end{algorithmic}
\end{algorithm}

The choice of MPC horizon $T$ has a significant effect on the performance of our approach. Since our latent variable encodes a skill which only partially completes the task, executing a single skill for too long unnecessarily penalizes a locally-useful skill for not being globally optimal. Hence, we set the MPC horizon $T$ to not more than twice the number of steps that a latent is actuated in the real environment $N$.

\section{EXPERIMENTS}
We evaluate our approach by completing two sequencing tasks on a Sawyer robot: drawing a sequence of points and pushing a box along a sequential path. For each of the experiments, the robot must complete an overall task by sequencing skills learned during the embedding learning process. Sequencing skills poses a challenge to conventional RL algorithms due to the sparsity of rewards in sequencing tasks~\cite{andrychowicz2017hindsight}. Because the agent only receives a reward for completing several correct complex actions in a row, exploration under these  sequencing tasks is very difficult for conventional RL algorithms. By reusing the skills we have consolidated in the embedding space, we show a high-level controller can effectively compose these skills in order to achieve such difficult sequencing tasks.

\subsection{Sawyer: Drawing a Sequence of Points}
\begin{figure}[H]
    \centering
    \includegraphics[width=\columnwidth]{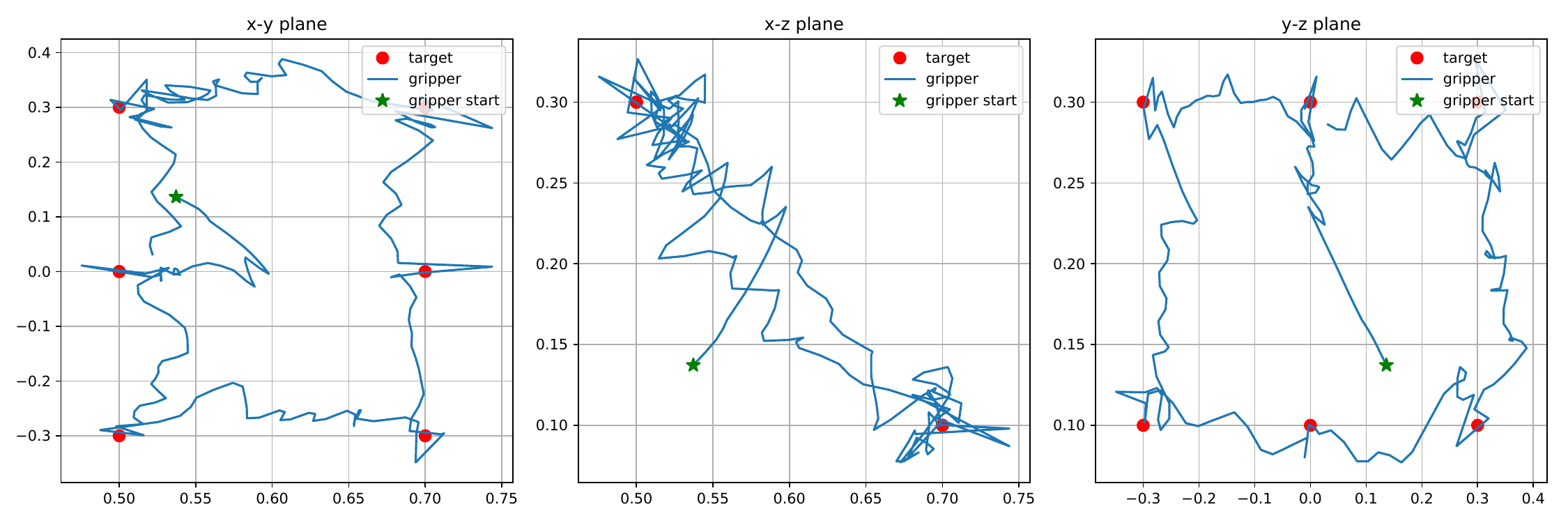}
    \caption{Gripper position plots for the unseen rectangle-drawing experiment in simulation. In this experiment, the unseen task is drawing a rectangle in 3D space.}
    \label{fig:reacher_sim_results_1}
\end{figure}

\begin{figure}[H]
    \centering
    \includegraphics[width=\columnwidth]{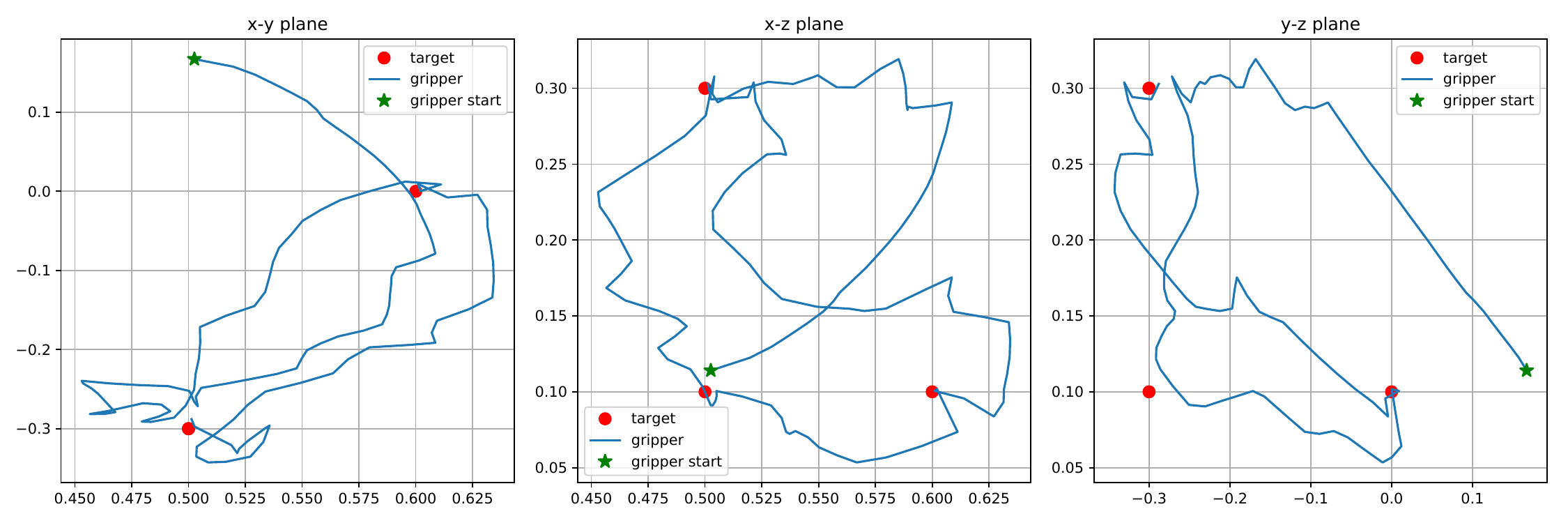}
    \caption{Gripper position plots in unseen triangle-drawing experiment in simulation. In this experiment, the unseen task is to move the gripper to draw a triangle.}
    \label{fig:reacher_sim_results_2}
\end{figure}

\begin{figure}[H]
    \centering
    \includegraphics[width=\columnwidth]{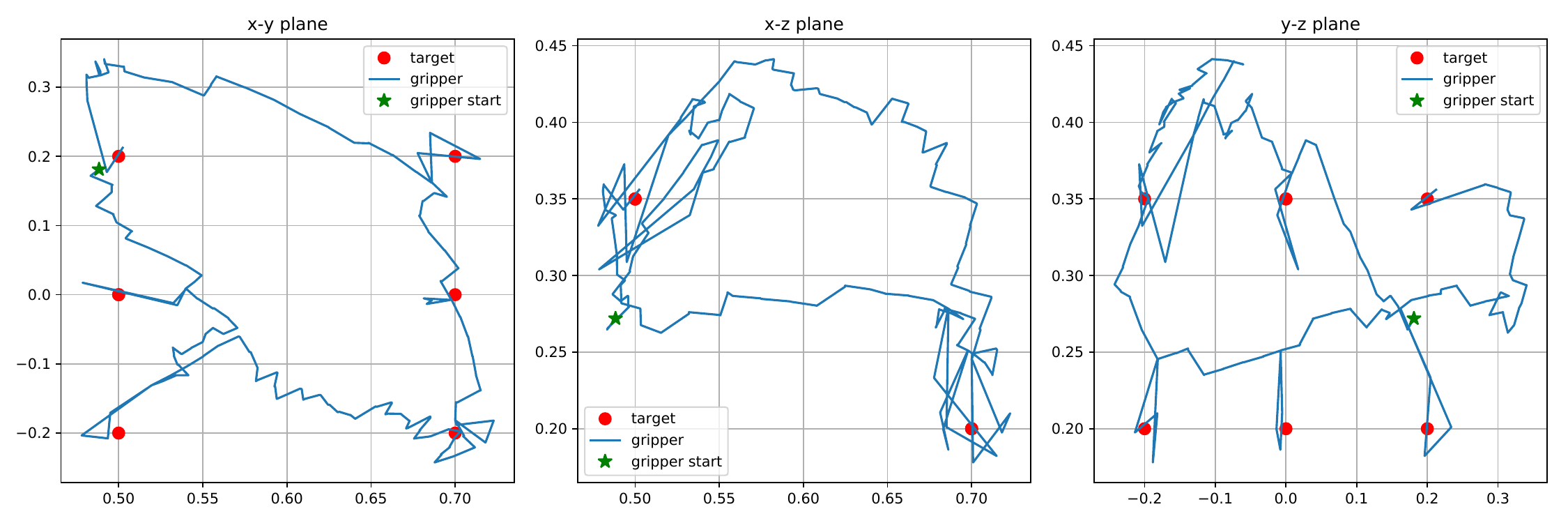}
    \caption{Gripper position plots for the triangle-drawing experiment on the real robot. In this experiment, the unseen task is to draw a triangle.}
    \label{fig:reacher_real_results_1}
\end{figure}

\begin{figure}[H]
    \centering
    \includegraphics[width=\columnwidth]{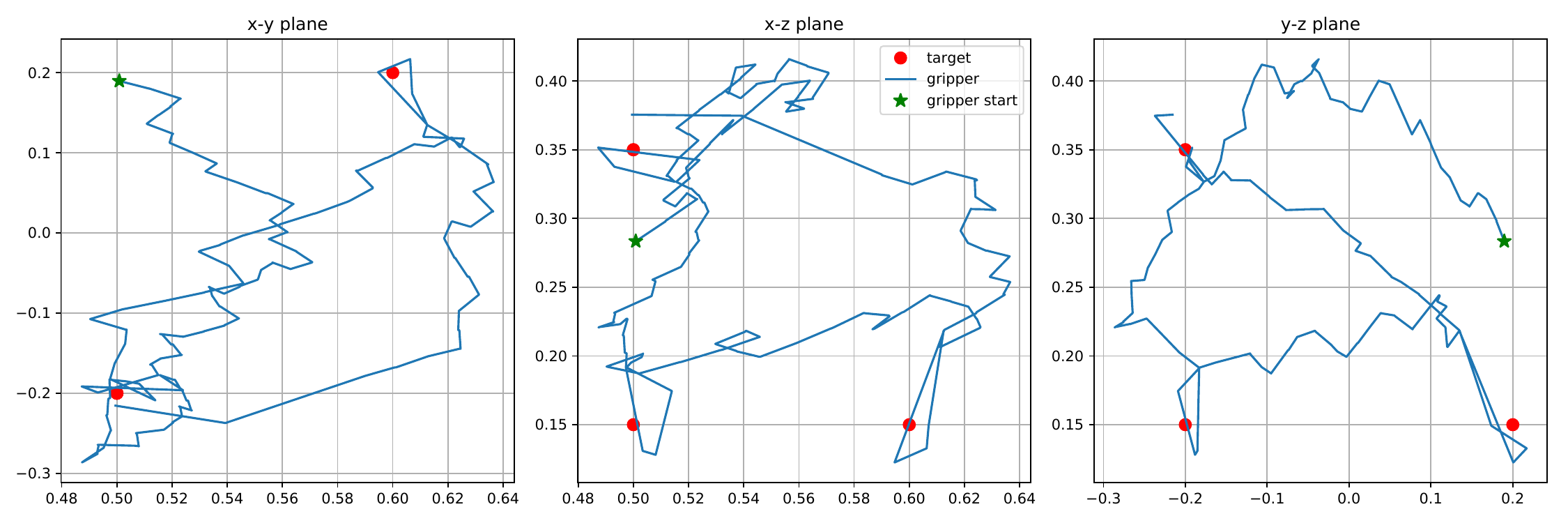}
    \caption{Gripper position plots in unseen triangle-drawing experiment on the real robot. In this experiment, the unseen task it to move the gripper to draw a triangle.}
    \label{fig:reacher_real_results_2}
\end{figure}

In this experiment, we ask the Sawyer Robot to move its end-effector to a sequence of points in 3D space. We first learn the low level policy that receives an observation with the robot's seven joint angles as well as the Cartesian position of the robot's gripper, and output incremental joint positions (up to 0.04 rads) as actions. We use the Euclidean distance between the gripper position and the current target is used as the cost function. We trained the policy and the embedding network on eight goal positions in simulation, forming a 3D rectoid enclosing the workspace. Then, we use the model-predictive control to choose a sequence latent vector which allows the robot to draw an unseen shape. For both simulation and real robot experiments, we attempted two unseen tasks: drawing a rectangle in 3D space (Figs.~\ref{fig:reacher_sim_results_1} and~\ref{fig:reacher_real_results_1}) and drawing a triangle in 3D space (Figs.~\ref{fig:reacher_sim_results_2} and~\ref{fig:reacher_real_results_2}).

\subsection{Sawyer: Pushing the Box through a Sequence of Waypoints}
In this experiment, we test our approach with a task that requires contact between the Sawyer Robot and an object. We ask the robot to push a box along a sequence of points in the table plane. We choose the Euclidean distance between the position of the box and the current target position as the reward function. The policy receives a state observation with the relative position vector between the robot's gripper and the box's centroid and outputs incremental gripper movements (up to $\pm$\SI{0.03}{\cm}) as actions.

We first pre-train a policy to push the box to four goal locations relative to its starting position in simulation. We trained the low-level multi-task policy with four tasks in simulation: \SI{20}{\cm} up, down, left, and right of the box starting position. We then use the model-predictive control to choose a latent vectors and feed it with the state observation to frozen multi-task policy which controls the robot.

For both simulation and real robot experiments, we use the simulation as a model of the environments. In the simulation experiments, we use model-predictive controller to push the box to three points. In the real robot experiments, we ask the Sawyer Robot to complete two unseen tasks: pushing up-then-left and pushing left-then-down.

\begin{figure}[H]
    \centering
    \includegraphics[width=\columnwidth]{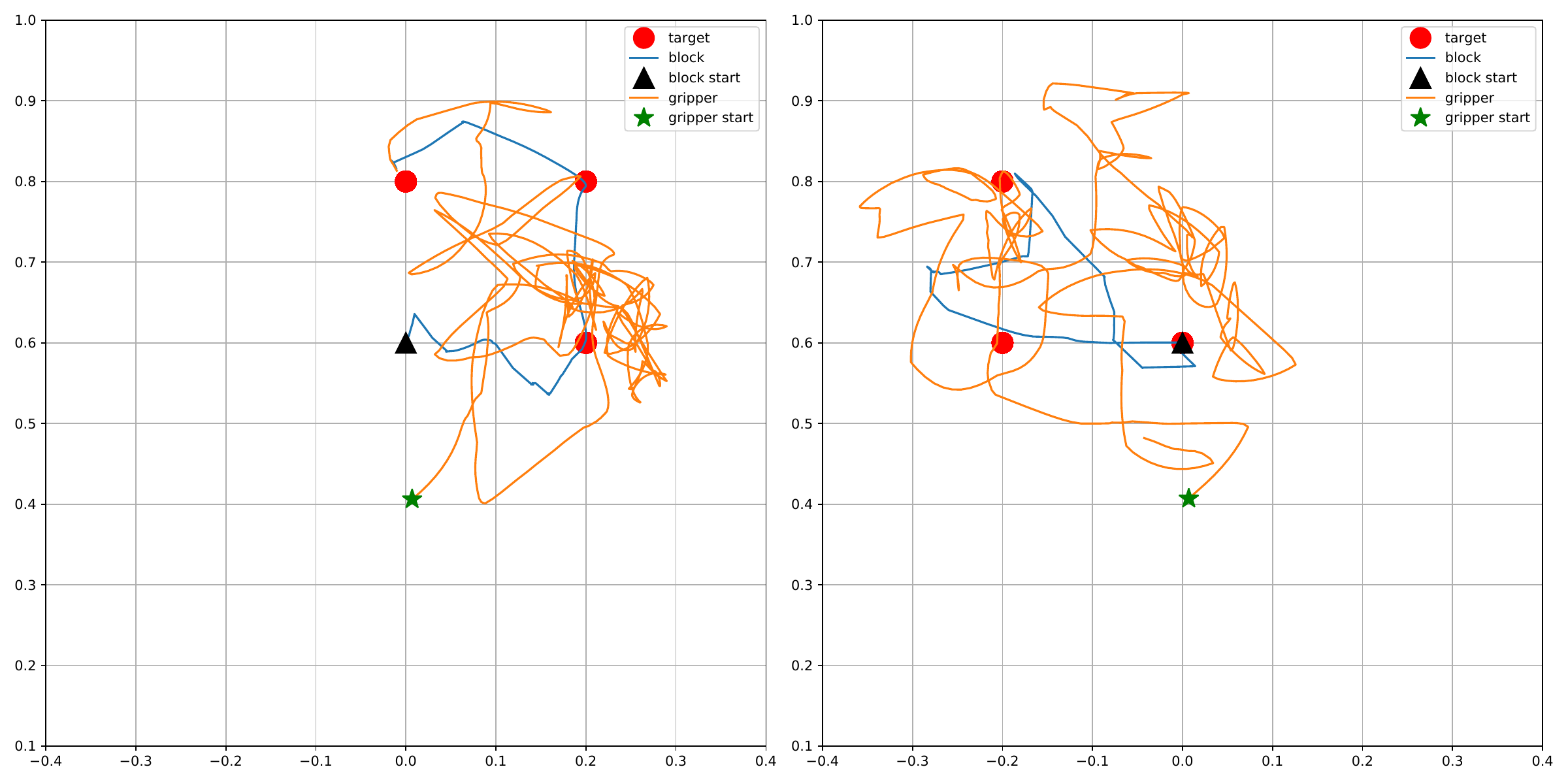}
    \caption{Plot of block positions and gripper positions in simulation experiments. In the first experiment (left), the robot pushes the box to the right, up and then left. In the second experiment (right), the robot pushes the box to the left, then up, and then back to its starting position.}
    \label{fig:pusher_sim_results}
\end{figure}

\begin{figure}[H]
    \centering
    \includegraphics[width=\columnwidth]{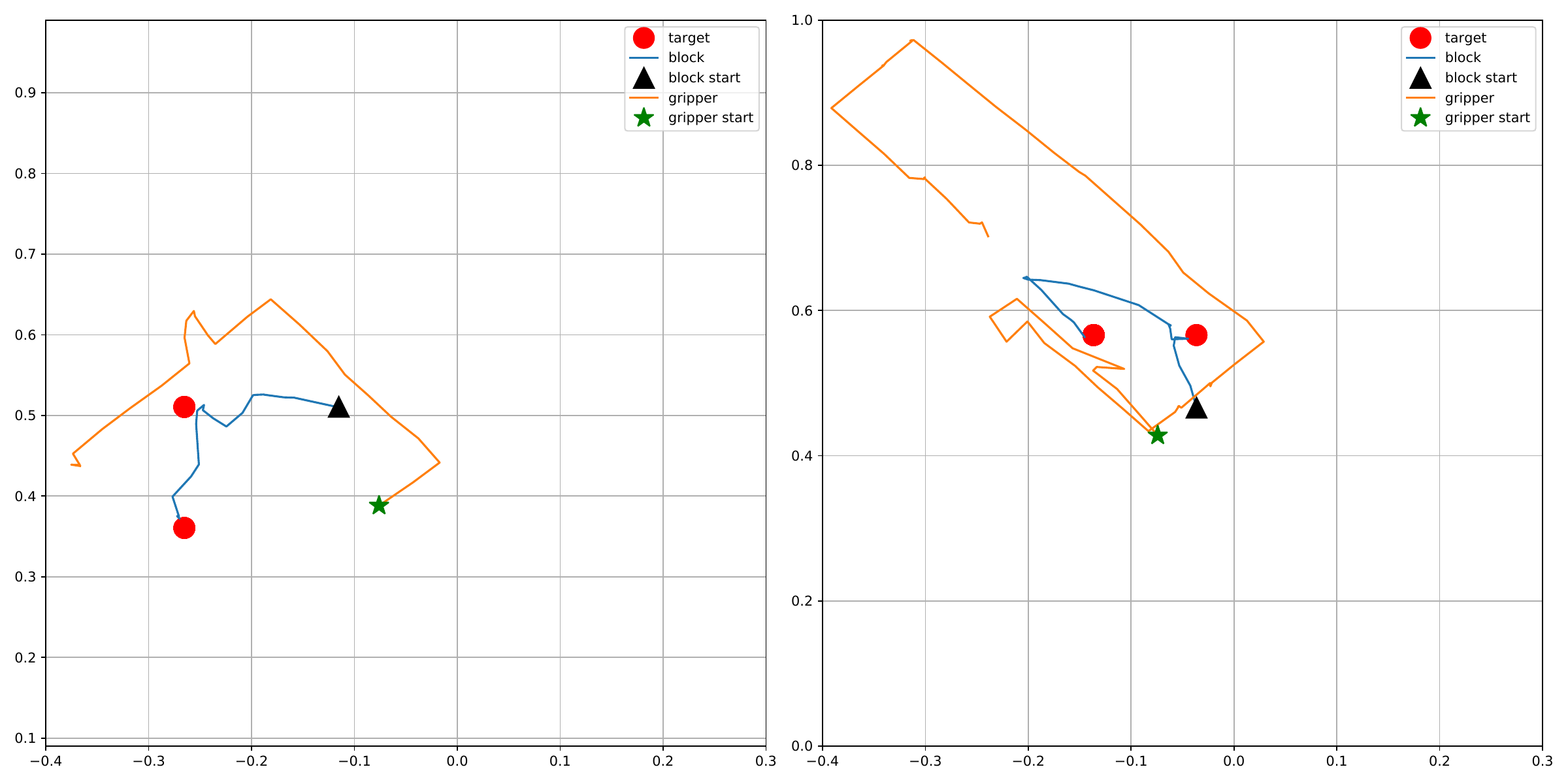}
    \caption{Plot of block positions and gripper positions in real robot experiments. In experiment I (left), the robot pushes the box to the left, and then down. In experiment II (right), the robot push the box to the up, and then left.}
    \label{fig:pusher_real_results}
\end{figure}

\section{RESULTS}

\subsection{Sawyer Drawing}

In the unseen drawing experiments, we sampled $k = 15$ vectors from the skill latent distribution, and for each of them performed an MPC optimization with a horizon of $T=4$ steps. We then execute the latent with highest reward for $N=2$ steps on the target robot. In simulation experiments, the Sawyer Robot successfully draw a rectangle with by sequencing 54 latents (Fig. 2) and drew by sequencing a triangle with 56 latents (Fig. 3). In the real robot experiments, the Sawyer Robot successfully completed the unseen rectangle-drawing task by choosing 62 latents (Fig. 4) in 2 minutes of real time and completed the unseen triangle-drawing task by choosing 53 latents (Fig. 5) in less than 2 minutes.

\subsection{Sawyer Pusher Sequencing}

In the pusher sequencing experiments, we sample $k = 50$ vectors from the latent distribution. We use an MPC optimization with a simulation horizon of $T=30$ steps, and execute each chosen latent in the environment for $N=10$ steps. In simulation experiments, the robot completed the unseen up-left task less than 30 seconds of equivalent real time and the unseen right-up-left task less than 40 seconds of equivalent real time. In the real robot experiments, the robot successfully completed the unseen left-down task by choosing 3 latents over approximately 1 minute of real time, and the unseen push up-left task by choosing 8 latents in about 1.5 minutes of real time.

\subsection{Analysis}

These experiment results show that our learned skills are composable to complete the new task. In comparison with performing a search as done in \cite{julian2018scaling}, our approach is faster in wall clock time because we perform the model prediction in simulation instead of on the real robot. Note that our approach can utilize the continuous space of latents, whereas previous search methods only use an artificial discretization of the continuous latent space. In the unseen box-pushing real robot experiment (Fig. 7, \textit{Right}), the Sawyer robot pushes the box towards the bottom-right right of the workspace to fix an error it made earlier in the task. This intelligent reactive behavior was never explicitly trained during the pre-training in simulation process. This shows that by sampling from our latent space, the model-predictive controller successfully selects a skill that is not pre-defined during training process.

\section{CONCLUSION}

In this work, we combine task representation learning simulation-to-real training, and model-predictive control to efficiently acquire policies for unseen tasks with no additional training. Our experiments show that applying model predictive control to these learned skill representations can be a very efficient method for online learning of tasks. The tasks we demonstrated are more complex than the underlying pre-trained skills used to achieve them, and the behaviors exhibited by our robot while executing unseen tasks were more adaptive than demanded by the simple reward functions us. Our method provides a partial escape from the reality gap problem in simulation-to-real methods, by mixing simulation-based long-range foresight with locally-correct online behavior.

For future work, we plan to apply our model-predictive controller as an exploration strategy to learn a composer policy that uses the latent space as action space. We look forward to efficiently learning a policy on real robots with guided exploration in our latent space.

\section*{ACKNOWLEDGEMENTS}
The authors would like to thank Angel Gonzalez Garcia, Jonathon Shen, and Chang Su for their work on the garage\footnote{https://github.com/rlworkgroup/garage} reinforcement learning for robotics framework, on which the software for this work was based.
We also want to thank the authors of multiworld\footnote{https://github.com/vitchyr/multiworld} for providing a well-tuned Sawyer Block Pushing simulation environment.
This research was supported in part by National Science Foundation grants IIS-1205249, IIS-1017134, EECS-0926052, the Office of Naval Research, the Okawa Foundation, and the Max-Planck-Society. Any opinions, findings, and conclusions or recommendations expressed in this material are those of the author(s) and do not necessarily reflect the views of the funding organizations.

\bibliography{literature}

\begin{thebibliography}{10}
\providecommand{\url}[1]{#1}
\csname url@samestyle\endcsname
\providecommand{\newblock}{\relax}
\providecommand{\bibinfo}[2]{#2}
\providecommand{\BIBentrySTDinterwordspacing}{\spaceskip=0pt\relax}
\providecommand{\BIBentryALTinterwordstretchfactor}{4}
\providecommand{\BIBentryALTinterwordspacing}{\spaceskip=\fontdimen2\font plus
\BIBentryALTinterwordstretchfactor\fontdimen3\font minus
  \fontdimen4\font\relax}
\providecommand{\BIBforeignlanguage}[2]{{%
\expandafter\ifx\csname l@#1\endcsname\relax
\typeout{** WARNING: IEEEtran.bst: No hyphenation pattern has been}%
\typeout{** loaded for the language `#1'. Using the pattern for}%
\typeout{** the default language instead.}%
\else
\language=\csname l@#1\endcsname
\fi
#2}}
\providecommand{\BIBdecl}{\relax}
\BIBdecl

\bibitem{mnih2015dqn}
\BIBentryALTinterwordspacing
V.~Mnih, K.~Kavukcuoglu, D.~Silver, A.~Rusu, J.~Veness, M.~Bellemare,
  A.~Graves, M.~Riedmiller, A.~Fidjeland, G.~Ostrovski, S.~Petersen,
  C.~Beattie, A.~Sadik, I.~Antonoglou, H.~King, D.~Kumaran, D.~Wierstra,
  S.~Legg, and D.~Hassabis, ``{Human-level control through deep reinforcement
  learning},'' \emph{Nature}, vol. 518, no. 7540, pp. 529--533, 2015. [Online].
  Available:
  \url{https://www.nature.com/nature/journal/v518/n7540/pdf/nature1423\\6.pdf}
\BIBentrySTDinterwordspacing

\bibitem{DBLP:journals/nature/SilverHMGSDSAPL16}
\BIBentryALTinterwordspacing
D.~Silver, A.~Huang, C.~J. Maddison, A.~Guez, L.~Sifre, G.~van~den Driessche,
  J.~Schrittwieser, I.~Antonoglou, V.~Panneershelvam, M.~Lanctot, S.~Dieleman,
  D.~Grewe, J.~Nham, N.~Kalchbrenner, I.~Sutskever, T.~P. Lillicrap, M.~Leach,
  K.~Kavukcuoglu, T.~Graepel, and D.~Hassabis, ``Mastering the game of go with
  deep neural networks and tree search,'' \emph{Nature}, vol. 529, no. 7587,
  pp. 484--489, 2016. [Online]. Available:
  \url{https://doi.org/10.1038/nature16961}
\BIBentrySTDinterwordspacing

\bibitem{DBLP:journals/corr/YahyaLKCL16}
\BIBentryALTinterwordspacing
A.~Yahya, A.~Li, M.~Kalakrishnan, Y.~Chebotar, and S.~Levine, ``Collective
  robot reinforcement learning with distributed asynchronous guided policy
  search,'' \emph{CoRR}, vol. abs/1610.00673, 2016. [Online]. Available:
  \url{http://arxiv.org/abs/1610.00673}
\BIBentrySTDinterwordspacing

\bibitem{10.1007/3-540-59496-5_337}
N.~Jakobi, P.~Husbands, and I.~Harvey, ``Noise and the reality gap: The use of
  simulation in evolutionary robotics,'' in \emph{Advances in Artificial Life},
  F.~Mor{\'a}n, A.~Moreno, J.~J. Merelo, and P.~Chac{\'o}n, Eds.\hskip 1em plus
  0.5em minus 0.4em\relax Berlin, Heidelberg: Springer Berlin Heidelberg, 1995,
  pp. 704--720.

\bibitem{closingrealitygap}
\BIBentryALTinterwordspacing
A.~A. Visser, N.~Dijkshoorn, M.~V.~D. Veen, and R.~Jurriaans, ``Closing the gap
  between simulation and reality in the sensor and motion models of an
  autonomous ar.drone.'' [Online]. Available:
  \url{http://citeseerx.ist.psu.edu/viewdoc/download?doi=10.1.1.841.1746\&re\\p=rep1\&type=pdf}
\BIBentrySTDinterwordspacing

\bibitem{peng2017simreal}
\BIBentryALTinterwordspacing
X.~Peng, M.~Andrychowicz, W.~Zaremba, and P.~Abbeel, ``Sim-to-real transfer of
  robotic control with dynamics randomization,'' \emph{CoRR}, vol.
  abs/1710.06537, 2017. [Online]. Available:
  \url{http://arxiv.org/abs/1710.06537}
\BIBentrySTDinterwordspacing

\bibitem{DBLP:journals/corr/SchulmanAC17}
\BIBentryALTinterwordspacing
J.~Schulman, P.~Abbeel, and X.~Chen, ``Equivalence between policy gradients and
  soft q-learning,'' \emph{CoRR}, vol. abs/1704.06440, 2017. [Online].
  Available: \url{http://arxiv.org/abs/1704.06440}
\BIBentrySTDinterwordspacing

\bibitem{DBLP:journals/corr/KingmaW13}
\BIBentryALTinterwordspacing
D.~P. Kingma and M.~Welling, ``Auto-encoding variational bayes,'' \emph{CoRR},
  vol. abs/1312.6114, 2013. [Online]. Available:
  \url{http://arxiv.org/abs/1312.6114}
\BIBentrySTDinterwordspacing

\bibitem{pastor2012asm}
P.~Pastor, M.~Kalakrishnan, L.~Righetti, and S.~Schaal, ``Towards associative
  skill memories,'' in \emph{Humanoids}, Nov 2012.

\bibitem{rueckert2015movprim}
E.~Rueckert, J.~Mundo, A.~Paraschos, J.~Peters, and G.~Neumann, ``Extracting
  low-dimensional control variables for movement primitives,'' in \emph{ICRA},
  May 2015.

\bibitem{hausman2018learning}
\BIBentryALTinterwordspacing
K.~Hausman, J.~Springenberg, Z.~Wang, N.~Heess, and M.~Riedmiller, ``Learning
  an embedding space for transferable robot skills,'' in \emph{ICLR}, 2018.
  [Online]. Available: \url{https://openreview.net/forum?id=rk07ZXZRb}
\BIBentrySTDinterwordspacing

\bibitem{julian2018scaling}
\BIBentryALTinterwordspacing
R.~C. Julian, E.~Heiden, Z.~He, H.~Zhang, S.~Schaal, J.~Lim, G.~S. Sukhatme,
  and K.~Hausman, ``Scaling simulation-to-real transfer by learning composable
  robot skills,'' in \emph{International Symposium on Experimental
  Robotics}.\hskip 1em plus 0.5em minus 0.4em\relax Springer, 2018. [Online].
  Available: \url{https://ryanjulian.me/iser\_2018.pdf}
\BIBentrySTDinterwordspacing

\bibitem{gu2017deep}
S.~Gu, E.~Holly, T.~Lillicrap, and S.~Levine, ``Deep reinforcement learning for
  robotic manipulation with asynchronous off-policy updates,'' in
  \emph{ICRA}.\hskip 1em plus 0.5em minus 0.4em\relax IEEE, 2017.

\bibitem{eysenbach2018diversity}
\BIBentryALTinterwordspacing
B.~Eysenbach, A.~Gupta, J.~Ibarz, and S.~Levine, ``Diversity is all you need:
  Learning skills without a reward function,'' Feb. 2018. [Online]. Available:
  \url{http://arxiv.org/abs/1802.06070}
\BIBentrySTDinterwordspacing

\bibitem{DBLP:journals/corr/KamtheD17}
\BIBentryALTinterwordspacing
S.~Kamthe and M.~P. Deisenroth, ``Data-efficient reinforcement learning with
  probabilistic model predictive control,'' \emph{CoRR}, vol. abs/1706.06491,
  2017. [Online]. Available: \url{http://arxiv.org/abs/1706.06491}
\BIBentrySTDinterwordspacing

\bibitem{sadeghi2017cadrl}
F.~Sadeghi and S.~Levine, ``{CAD2RL}: Real single-image flight without a single
  real image,'' in \emph{RSS}, 2017.

\bibitem{AAMASWS10-barrett}
\BIBentryALTinterwordspacing
S.~Barrett, M.~E. Taylor, and P.~Stone, ``Transfer learning for reinforcement
  learning on a physical robot,'' in \emph{Ninth International Conference on
  Autonomous Agents and Multiagent Systems - Adaptive Learning Agents Workshop
  (AAMAS - ALA)}, May 2010. [Online]. Available:
  \url{http://www.cs.utexas.edu/users/ai-lab/?AAMASWS10-barrett}
\BIBentrySTDinterwordspacing

\bibitem{DBLP:journals/corr/abs-1803-07551}
\BIBentryALTinterwordspacing
S.~S{\ae}mundsson, K.~Hofmann, and M.~P. Deisenroth, ``Meta reinforcement
  learning with latent variable gaussian processes,'' \emph{CoRR}, vol.
  abs/1803.07551, 2018. [Online]. Available:
  \url{http://arxiv.org/abs/1803.07551}
\BIBentrySTDinterwordspacing

\bibitem{coreyes2018self}
\BIBentryALTinterwordspacing
J.~D. Co-Reyes, Y.~Liu, A.~Gupta, B.~Eysenbach, P.~Abbeel, and S.~Levine,
  ``{Self-Consistent} trajectory autoencoder: Hierarchical reinforcement
  learning with trajectory embeddings,'' Jun. 2018. [Online]. Available:
  \url{http://arxiv.org/abs/1806.02813}
\BIBentrySTDinterwordspacing

\bibitem{schulman2017ppo}
\BIBentryALTinterwordspacing
J.~Schulman, F.~Wolski, P.~Dhariwal, A.~Radford, and O.~Klimov, ``Proximal
  policy optimization algorithms,'' \emph{CoRR}, vol. abs/1707.06347, 2017.
  [Online]. Available: \url{http://arxiv.org/abs/1707.06347}
\BIBentrySTDinterwordspacing

\bibitem{todorov2012mujoco}
``{MuJoCo: A physics engine for model-based control},'' in \emph{IROS}, 2012.

\bibitem{andrychowicz2017hindsight}
M.~Andrychowicz, F.~Wolski, A.~Ray, J.~Schneider, R.~Fong, P.~Welinder,
  B.~McGrew, J.~Tobin, P.~Abbeel, and W.~Zaremba, ``Hindsight experience
  replay,'' in \emph{NIPS}, 2017.

\end{thebibliography}
\bibliographystyle{IEEEtran}

\end{document}